\documentclass[a4paper,conference]{IEEEtran}
\ifCLASSINFOpdf
   \usepackage[pdftex]{graphicx}
\else
\fi
\usepackage{amsmath,amssymb,amsfonts}
\usepackage{pifont}
\newcommand{\cmark}{\ding{51}}%
\newcommand{\xmark}{\ding{55}}%
\usepackage{array}
\begin{document}
%
\title{Temporal Attention-Augmented Graph Convolutional Network for Efficient Skeleton-Based Human Action Recognition}


\author{\IEEEauthorblockN{Negar Heidari and Alexandros Iosifidis}\\
\IEEEauthorblockA{Department of Electrical and Computer Engineering, Aarhus University, Denmark\\
Emails: \{negar.heidari, ai\}@ece.au.dk}}


%


\maketitle

\begin{abstract}
Graph convolutional networks (GCNs) have been very successful in modeling non-Euclidean data structures, like sequences of body skeletons forming actions modeled as spatio-temporal graphs. Most GCN-based action recognition methods use deep feed-forward networks with high computational complexity to process all skeletons in an action. This leads to a high number of floating point operations (ranging from 16G to 100G FLOPs) to process a single sample, making their adoption in restricted computation application scenarios infeasible. 
In this paper, we propose a temporal attention module (TAM) for increasing the efficiency in skeleton-based action recognition by selecting the most informative skeletons of an action at the early layers of the network. We incorporate the TAM in a light-weight GCN topology to further reduce the overall number of computations. 
Experimental results on two benchmark datasets show that the proposed method outperforms with a large margin the baseline GCN-based method while having $\times2.9$ less number of computations. Moreover, it performs on par with the state-of-the-art with up to $\times9.6$ less number of computations. 
\end{abstract}


%
\IEEEpeerreviewmaketitle

\section{Introduction}
Human action recognition has been a very popular research topic in recent years. RGB videos and different types of modalities such as depth, optical flow and human body skeletons can be used for this task \cite{du2015hierarchical,liu2016spatio,iosifidis2012view}. Compared to other data modalities, human body skeletons encode compact and high level information representing the human pose and joints’ motion while being invariant to viewpoint variations, motion speed, human appearance, and body scale \cite{han2017space}, and it is robust to context noise. 
Considering the success of existing pose estimation techniques \cite{shotton2011real,sun2019deep,cao2017realtime} and the availability of depth cameras \cite{zhang2012microsoft}, obtaining the human body skeleton data is much easier than before. 
Thus, skeleton-based human action recognition has attracted an increasing research interest in recent years and many deep learning-based methods were proposed to model both the spatial and temporal evolution of skeletons in a sequence. Some methods use recurrent neural network (RNN) architectures, like the Long Short-Term Memory (LSTM) network \cite{greff2016lstm}, which are suitable to model temporal dynamics \cite{du2015hierarchical,liu2016spatio,shahroudy2016ntu,song2017end,zhang2017view,li2018skeleton}
and utilize the skeleton data as vectors formed by the body joints coordinates. Methods employing Convolutional Neural Networks (CNN) \cite{liu2017two,kim2017interpretable,ke2017new,liu2017enhanced,li2017skeleton,li2017skeletonCNN} reorganize the body joints' coordinates of each pose to a 2D map which is a suitable input format for CNNs. The high model complexity in all these methods make their training and inference processes very time consuming \cite{ke2017new,pascanu2014construct}.
Besides, these methods are not able to completely benefit from the non-euclidean structure of the skeleton data which represents the spatial body joints connections.

Graph Convolutional Networks (GCNs) have been very successful when applied to many pattern recognition tasks \cite{kipf2016semi,duvenaud2015convolutional,niepert2016learning,atwood2016diffusion,hamilton2017inductive,monti2017geometric,kipf2018neural,heidari2020progressive}, by generalizing the convolution operation from grid data into graph data structures. Recently, significant results have been obtained by employing GCNs for skeleton-based human action recognition \cite{shi2019two,yan2018spatial,peng2020learning,tang2018deep,li2019actional,shi2019skeleton_directed,gao2019optimized}. In these methods, an action is represented as a sequence of body poses and each body pose is represented by a skeleton. The skeleton data is treated as a graph which models the spatial relationship between different body joints and the temporal dynamics in an action are expressed by a sequence of skeletons. 

Most of the recently proposed GCN-based methods use deep feed forward networks to model the spatio-temporal features of body skeletons. Considering that these methods process all the body sekeletons in a sequence depicting the performed action, this approach is not efficient in terms of memory consumption and computation time. While memory efficiency can be increased by employing network compression, weight prunning, quantization and low-rank approximation approaches \cite{zhang2019real,courbariaux2015binaryconnect,han2015deep,tran2018improving}, by processing a large number of body skeletons in each sequence, the number of floating point operations (FLOPs) is still large. Thus, to address both memory and computational efficiency in skeleton-based human action recognition, not only we need more compact and lightweight network architectures, but also the number of FLOPs should be minimized. Hence, processing fewer body skeletons for action recognition is a large step towards increasing the computational efficiency of both training and inference processes. 

In this paper, we argue that all body skeletons in a temporal sequence are not equally important for recognizing actions. For each action class, there exist body poses which are the most informative for the action, and we can extract sufficient information for action recognition by focusing on the skeletons of these body poses only. 
Our goal is to increase computational efficiency while performing on par, or even better, compared to models which utilize all the body skeletons in a sequence for action recognition. In this regard, we propose a GCN-based model which is capable to select a subset of body skeletons for human action recognition. 
To select the most informative skeletons, we propose a trainable temporal attention module which measures the importance of each skeleton in a sequence and we employ this module in the GCN-based spatio-temporal model to increase its efficiency. 
The main contributions of our work are the following: 
\begin{itemize}
    \item We propose a temporal attention module (TAM) to extract the most informative skeletons in an action sequence, leading to increased computational efficiency in both the training phase and inference.
    \item We experimentally show that employing our proposed TAM with a light-weight GCN topology leads to state-of-the-art performance levels in two widely adopted datasets for skeleton-based action recognition.
    \item We show that a subset of skeletons is as informative as the full skeleton sequence, since our method performs on par with the state-of-the-art methods while increasing computational efficiency by a factor of up to $\times 9.6$. This is a major advantage of our method compared to the state-of-the-art, as it is suitable for real time action recognition under restricted computation scenarios.
\end{itemize}

The remainder of the paper is organized as follows. Section \ref{sec:related_work} discusses the related works and section \ref{sec:ST-GCN} introduces the baseline method ST-GCN \cite{yan2018spatial}. Section \ref{sec:proposed} describes the proposed method. The conducted experiments and results are presented in section \ref{sec:results}, and the concluding remarks are drawn in section \ref{sec:conclusion}. 

\section{Related work}
\label{sec:related_work}
Data driven skeleton-based human action recognition methods which use deep learning models are mainly categorized into RNN-based, CNN-based and GCN-based methods. RNN-based methods \cite{du2015hierarchical,liu2016spatio,shahroudy2016ntu,song2017end,zhang2017view,li2018skeleton} mostly employ LSTM networks \cite{greff2016lstm} to model the temporal dynamics of skeletons' sequence. In these methods, the sequence of skeletons is introduced to the model as a sequence of vectors. Each vector is formed by the concatenated 3D coordinates of all body joints of skeleton. ST-LSTM \cite{liu2016spatio} is one representative RNN-based method which captures the body joints' relationship in both temporal and spatial domain.
CNN-based methods \cite{chen2018cascaded,liu2017two,kim2017interpretable,ke2017new,liu2017enhanced,li2017skeleton,li2017skeletonCNN} convert the sequence of skeletons into pseudo-images by reorganizing the joints' coordinates into a 2D map and employ a state-of-the-art CNN model like ResNet \cite{he2016deep,szegedy2017inception} to extract temporal and spatial features. Although the CNN-based methods are easier to train than RNN-based methods, they employ deep network architectures with large receptive fields to perceive the semantics of the input map. 
Besides, since all these RNN-based and CNN-based methods convert the skeletons into a regular grid or a sequence, they cannot utilize the complex, irregular and non-Euclidean structure of the skeleton data. 

Recently, many GCN-based methods for skeleton-based human action recognition have been proposed which treat the human skeleton data as a graph structure representing the body joints (graph nodes) and their natural connections with bones (graph edges). Since the GCNs are able to capture the embedded features in irregular structured data, the GCN-based methods achieve higher performance in skeleton-based human action recognition compared to RNN-based and CNN-based methods. 
Spatio-temporal graph convolutional network (ST-GCN) \cite{yan2018spatial} is the first GCN-based method proposed for action recognition. It receives the sequence of body skeletons directly as input and employs the GCNs' \cite{kipf2016semi} aggregation rule to extract the spatial features of each skeleton in a sequence, while the temporal dynamics are modeled by using a temporal graph with fixed connections. 
Several methods have been proposed based on ST-GCN for skeleton based human action recognition which mostly focus on exploiting adaptive spatial graphs.  
2s-AGCN \cite{shi2019two} is one of the state-of-the-art methods proposed on top of ST-GCN \cite{yan2018spatial}. This method adaptively learns the topology of the graph in different layers for each sequence of skeletons in an end-to-end manner. It updates the graph structure of the skeleton with a spatial attention and also a data dependent graph. Besides, it uses a two-stream framework to utilize both joints features and bones features in parallel. It introduces joints and bones data into two different models and, finally, the SoftMax scores of the two model are added to obtain the fused score and predict the action label.
In DGNN \cite{shi2019skeleton_directed} the skeleton data is represented as a directed acyclic graph to benefit from the relationship between joints and bones based on their kinematic dependency and it improved 2s-AGCN \cite{shi2019two} by utilizing the motion data and also making the graph structure adaptive.
GCN-NAS \cite{peng2020learning} follows a neural architecture search approach to explore an optimal GCN-based model in terms of graph structures. That is, it explores the search space to determine the best graph structure at each layer of the network, while it still employs the ST-GCN network topology.
DPRL+GCNN \cite{tang2018deep} aims to select the most representative skeletons from the input sequence using deep reinforcement learning. It adjusts the chosen skeletons progressively by evaluating the models' performance in action recognition. 
Actional-Structural GCN (AS-GCN) \cite{li2019actional} proposed an inference module based on encoder-decoder structure to capture richer action-specific correlations and also structural links between the joints in skeletons.

\section{Spatial Temporal Graph Convolutional Network (ST-GCN)}
\label{sec:ST-GCN}
In this section, we describe ST-GCN model \cite{yan2018spatial} as our baseline. It is formulated over a sequence of skeleton graphs with multiple layers of spatio-temporal graph convolution operations. The model receives the joint coordinate vectors and the constructed spatio-temporal graph as input and applies GCN to capture the embedded patterns of the graph and extract high level features of the skeletons. Each convolution layer integrates the human body information along both the spatial and temporal dimensions. The extracted features are then passed to a fully connected layer which classifies the skeleton sequence to the action class using a standard SoftMax classifier. The entire model is trained with backpropagation to minimize the classification loss. 
The network architecture of ST-GCN \cite{yan2018spatial} is constructed by 9 layers of spatio-temporal convolution units. In each layer the Resnet module is also applied and the output is followed by a batch normalization layer. In the following, the graph construction, and the spatial and temporal convolution units are described. 

\subsection{Graph Construction}\label{subsec:GC}
\begin{figure}
    \centering
    \includegraphics[width=0.3\linewidth]{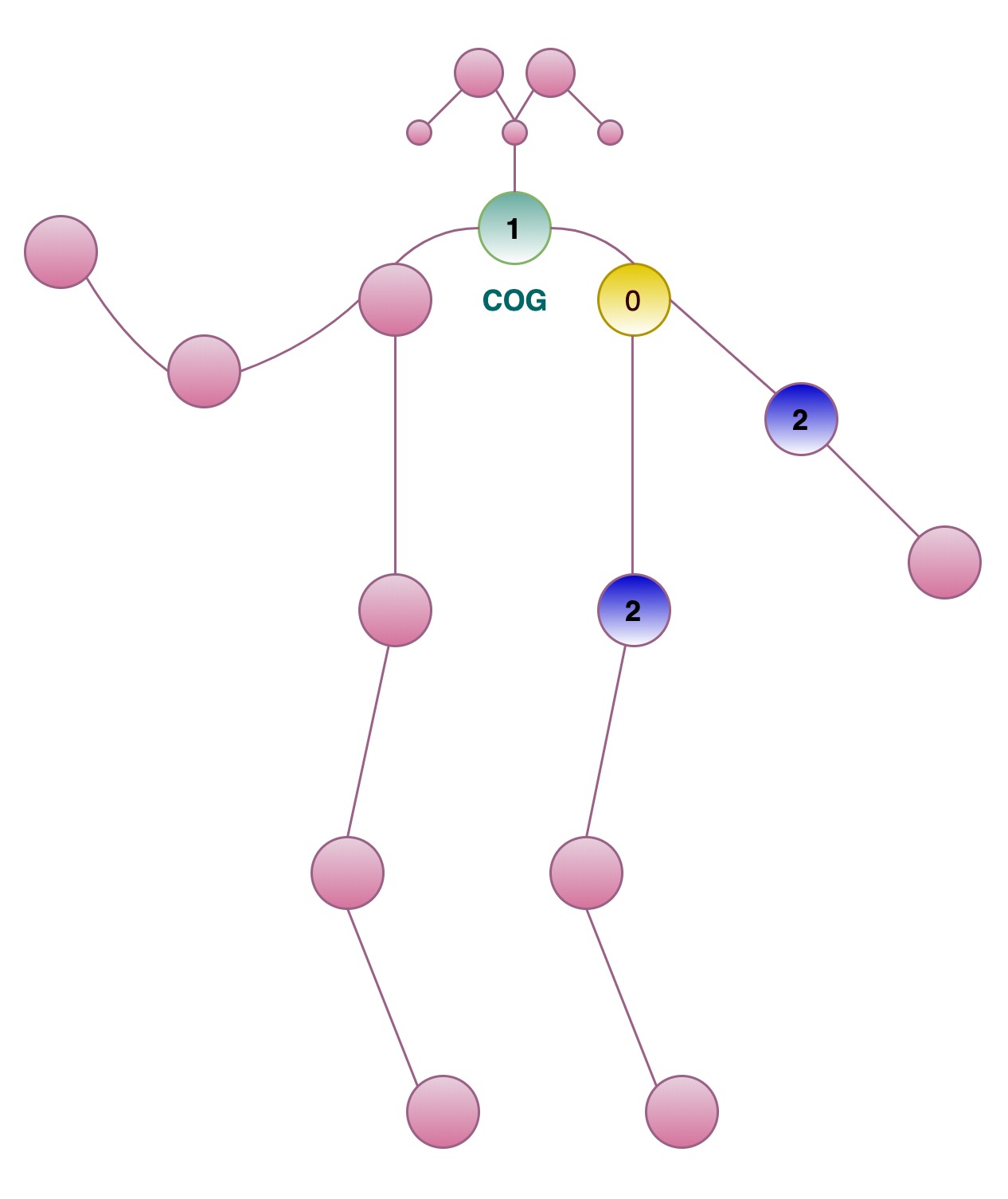}
    \includegraphics[width=0.4\linewidth]{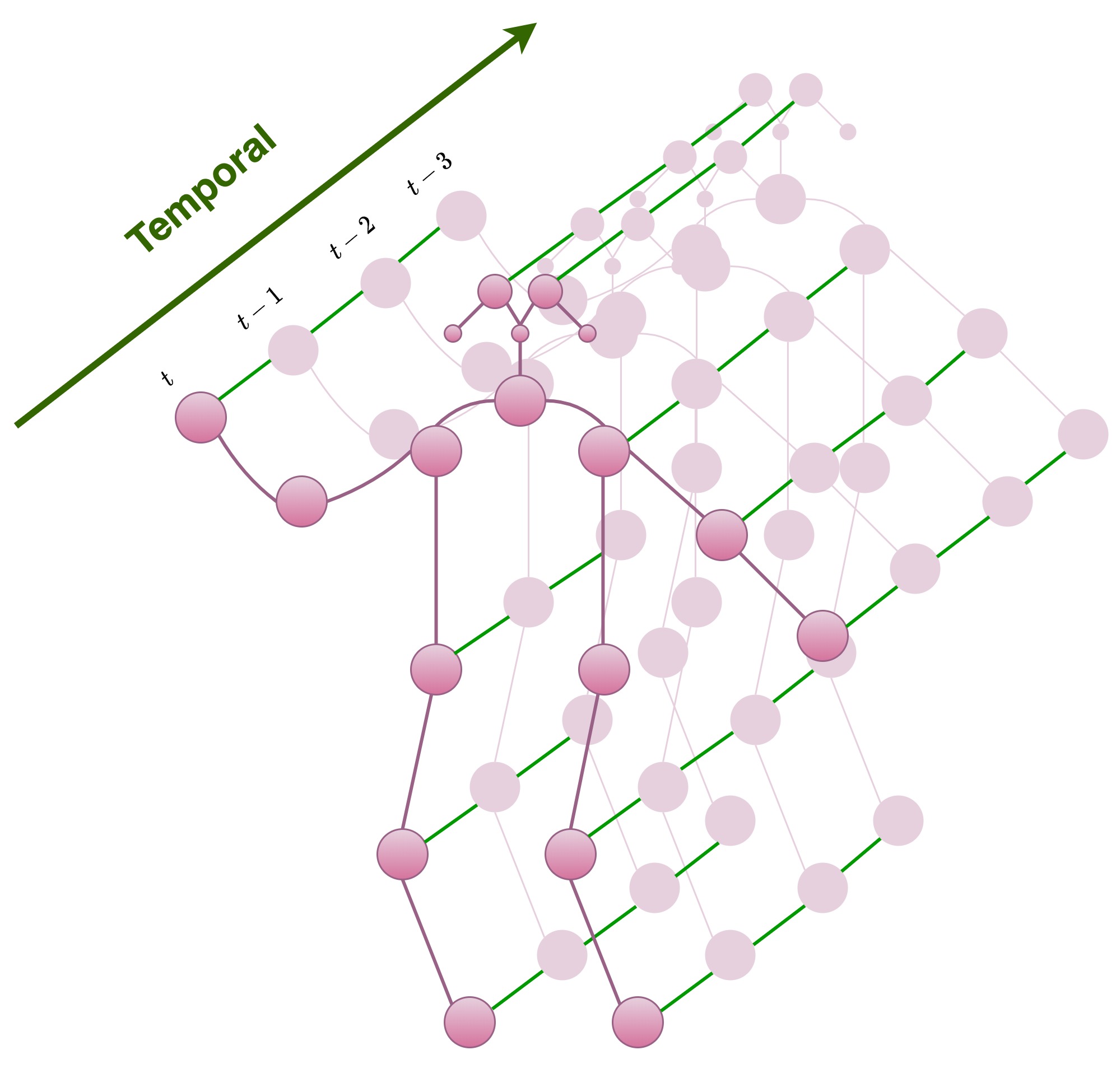}
    \caption{Illustration of Spatio-temporal graph (right) in TA-GCN, and the spatially partitioned skeleton (left). Each partition set is illustrated in a different color.}
    \label{fig:ST-Graph-Partitioning}
\end{figure}

Each skeleton is represented as a graph which forms the hierarchical representation of skeleton sequences and the spatial patterns and temporal dynamics of these graphs are captured automatically.
The spatio-temporal graph is constructed using the 2D or 3D human body joint coordinates as nodes, and spatial and temporal edges. The spatial edges correspond to the natural connectivity of body joints in the skeleton, and the temporal edges connect the same body joints across consecutive skeletons. Fig. \ref{fig:ST-Graph-Partitioning} (right) shows the spatio-temporal skeleton graph.  

Formally, an undirected spatio-temporal graph on a sequence of skeletons is denoted as $\mathcal{G} = (\mathcal{V} ,\mathcal{E})$, where the set of nodes:
\begin{equation}
\mathcal{V} = \left \{\nu_{ti}\mid t, \:i \in \mathbb{Z}, \:\:1\leq t\leq T, \:\:1\leq i\leq N\right \} \nonumber
\end{equation}
indicates $N$ body joints of a skeleton in a sequence of $T$ time steps and $\mathcal{E}$ is the set of spatial (intra-skeleton) and temporal (inter-skeleton) connections. The graph structure is captured by the adjacency matrix $\mathbf{A} \in \mathbb{R}^{N \times N}$ which is a symmetric binary matrix defined as $\mathbf{A}_{ij} = 1 $ if there is a connection between nodes $\nu_{ti}$ and $\nu_{tj}$ in time step $t$, otherwise $\mathbf{A}_{ij} = 0$. 
Given $\mathbf{A}$ as the spatial graph adjacency matrix which represents the joint connections in a single skeleton, the normalized adjacency matrix $\hat{\mathbf{A}}$ with self connections is computed as:
\begin{equation}
    \mathbf{\hat{A}} = \mathbf{\Tilde{D}}^{-\frac{1}{2}} \mathbf{\Tilde{A}} \mathbf{\Tilde{D}}^{-\frac{1}{2}}, 
    \label{eq:nomalized-A}
\end{equation}
where $\mathbf{\Tilde{A}} = \mathbf{A} + \mathbf{I}_{N}$ and $\mathbf{\Tilde{D}}$ is the diagonal degree matrix of $\mathbf{\Tilde{A}}$, i.e. $\mathbf{\Tilde{D}}_{{ii}} = \sum_{j}^{N} \mathbf{\Tilde{A}}_{{ij}}$.

The body motions are grouped as concentric and eccentric, the spatially localized structure of a single skeleton is used to design the spatial partitioning process which divides the neighbor nodes of each body joint into three subsets: 1) the root node itself; 2) the nodes connected to the root which are closer to the skeletons' center of gravity than the root node; 3) the remaining roots' neighbors. 
Fig. \ref{fig:ST-Graph-Partitioning} (left) shows the neighboring sets of each graph node which are presented with different colors \cite{yan2018spatial,shi2019two}. 

\subsection{Spatial Graph Convolution}
Given the input spatial features of $i^{th}$ joint, $f_{in}(\nu_{i})$, the spatial graph convolution is defined as \cite{yan2018spatial}: 
\begin{equation}
    f_{out}(\nu_{i}) =  \sum_{\nu_{j}\in \mathcal{N}_i} \frac{1}{Z_{ij}}f_{in}(\nu_{j})w(l_i(\nu_{j})), 
    \label{eq:conv_s}
\end{equation}
where $\mathcal{N}_i$ is the set of intra-frame neighbor vertices of $\nu_{i}$ which are directly connected to it. 
As mentioned in \ref{subsec:GC}, ST-GCN uses a partitioning method to divide the neighbor vertices of each node into three subsets. According to this partitioning process, $l_i$ in (\ref{eq:conv_s}) maps each neighbor vertex into one of these three subsets and $w$ is the weight matrix which is unique for each neighbor set and maps the target nodes' features into a new subspace. Note that the number of neighbor subsets and weight matrices are fixed while the number of neighbor nodes in each subset is varied for each node. $Z_{ij}$ is a normalization factor which balances the contribution of each set of neighbor nodes in producing $f_{out}(\nu_{i})$ which contains the output features of the target node $\nu_{i}$. 

In terms of implementation, the $C$-dimensional input feature vector for the node $\nu_{i}$ is denoted as $\mathbf{x}_{i} = f_{in}(\nu_{i}) $ and $\mathbf{X} \in \mathbb{R}^{C_{in} \times T \times N}$ represents the input feature tensor for a sequence of skeletons, where $C_{in}$ denotes the number of input channels, $T$ is the number of skeletons and $N$ is the number of body joints in each skeleton.  
The model receives the feature tensor $\mathbf{X}$ as input and updates the nodes' feature vectors by applying the spatial convolution to produce $\mathbf{X}^{\prime}\in \mathbb{R}^{C_{out}\times T\times N}$ with $C_{out}$ channels as output. By employing the layer-wise propagation rule of GCNs proposed in \cite{kipf2016semi}, the spatial convolution is formally defined as \cite{yan2018spatial}: 
\begin{equation}
    \mathbf{X}^{\prime} = ReLU\left(\sum_{p} (\mathbf{\hat{A}}_p\otimes \mathbf{M}_p)\mathbf{X}\mathbf{W}_p \right),
    \label{eq:2dConv_s}
\end{equation}
where $\otimes$ is the element-wise product of two matrices. 
\begin{figure*}
    \centering
    \includegraphics[width=1\textwidth]{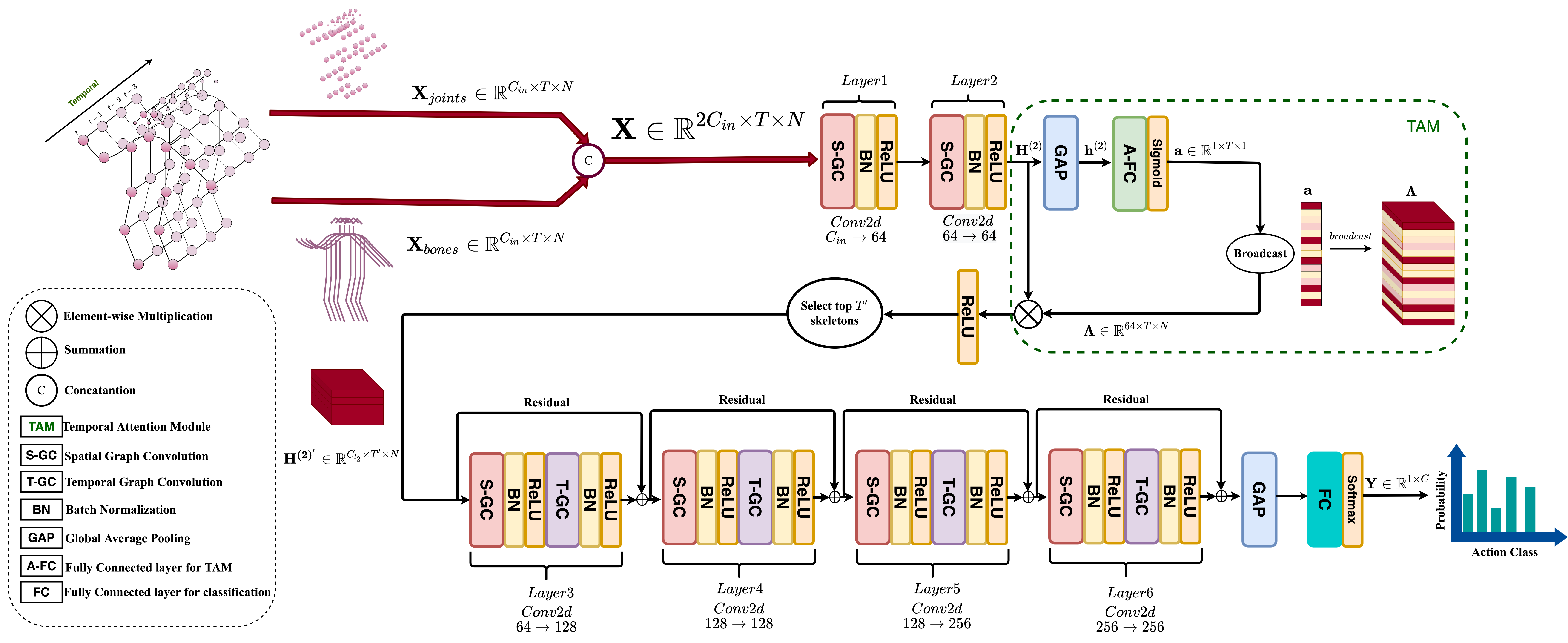}
    \caption{Illustration of the proposed model diagram. The whole model architecture is composed of 6 spatio-temporal layers and the temporal attention module places after the first two layers to select the most informative skeletons. The highlighted skeltons' indices are sorted in descending manner and then extracted from the $\mathbf{H}^{(2)}$. The $3^{rd}$ to $6^{th}$ layers employ both spatial and temporal convolution.}
    \label{fig:Model-Diagram}
\end{figure*}
According to the partitioning process described in section \ref{subsec:GC}, each node has 3 subsets of neighbors. Therefore, the ajdacency matrix $\mathbf{\Tilde{A}}$ is defined as the summation of $3$ different adjacency matrices which are indexed by $p$ as follows:
\begin{equation}
    \mathbf{\Tilde{A}} = \mathbf{A} + \mathbf{I}_N = \sum_{p}\mathbf{A}_p
\end{equation}
$\mathbf{A}_0 = \mathbf{I}_N$ represents the nodes' self connections and the normalized adjacency matrix for each subset is defined as: 
\begin{equation}
    \hat{\mathbf{A}}_p = \mathbf{D}^{-\frac{1}{2}}_p\mathbf{A}_p\mathbf{D}^{-\frac{1}{2}}_p,  
\end{equation}
where $\mathbf{D}_{({ii})_p} = \sum_{j}^{N} \mathbf{A}_{({ij})_p} + \varepsilon$, with $\varepsilon = 0.001$ used to avoid empty rows in degree matrix. 
$\mathbf{M}_p \in \mathbb{R}^{N \times N}$ is a learnable attention map which highlights the elements of each adjacency matrix and it is initialized as an all-one matrix. $\mathbf{W}_p \in \mathbb{R}^{C_{out} \times C_{in}}$ denotes the weight matrix which transforms the node features of each partition. In practice, a standard 2D convolution is used which performs $C_{out} \times 1 \times 1$ convolutions, then the resulting tensor is multiplied with the masked normalized adjacency matrix on the last dimension $N$.

\subsection{Temporal Graph Convolution}
To capture the temporal dynamics in a skeleton sequence and consider the motions taking place in an action, we need to propagate information related to the nodes' features in other skeletons of the sequence. To this end, a temporal convolution is applied on the output tensor of the spatial convolution step. In temporal dimension, each node is only connected to its corresponding node in its previous and next skeletons, so the number of neighbors for each node is fixed to 2. Therefore, a 2D convolution with a predefined temporal kernel size $K_t$ is is applied to the output of (\ref{eq:2dConv_s}), $\mathbf{X^{\prime}}$, to aggregate the features of each body joint in different time steps.

\section{Temporal Attention-Augmented Graph Convolutional Network (TA-GCN)}\label{sec:proposed}
In this section we describe an end-to-end temporal attention-augmented GCN model for efficient skeleton-based human action recognition. 
In order to extract discriminative features in temporal dimension of data, we propose TAM which highlights the most informative skeletons in a sequence. Based on this process, we apply skeleton selection to highly reduce the overall computational cost both during training and inference.

Similar to most of the recently proposed works \cite{yan2018spatial,shi2019two}, the proposed model is composed of multiple layers of spatio-temporal units. In each unit, the embedded spatial features in each skeleton are first extracted by employing a standard 2D convolution as described in Eq. (\ref{eq:2dConv_s}).
Motivated by \cite{shi2019two}, the learnable spatial attention map $M_p$ is added (and not element-wise multiplied) to the graph adjacency matrix $A_p$. Adding the attention map to the graph adjacency matrix has the advantage that, not only the strength of the existing graph edges can be highlighted, but relationships between graph nodes which are not connected in the adjacency matrix based on the physical connectivity of the human body skeleton joints can also be captured \cite{shi2019two}. Such connections can be important in effectively describing actions, e.g. encoding the relationships between the joints of the human arms and the head is particularly important for describing actions like `hand waving'. Thus, the spatial attention module is employed to selectively focus on the most informative joints in each skeleton and the spatial convolution in Eq. (\ref{eq:2dConv_s}) takes the following form: 
\begin{equation}
    \mathbf{X}^{\prime} = ReLU \left(\sum_{p} (\hat{\mathbf{A}}_p + \mathbf{M}_p)\mathbf{X}\mathbf{W}_p \right)
    \label{eq:Attn_Conv_s}
\end{equation}

The TAM takes a tensor $\mathbf{H}^{(l)} \in \mathbb{R}^{C_{l} \times T \times N}$ as input, which can be the input data, i.e. $\mathbf{H}^{(0)} = \mathbf{X}$, or the output of the $l^{th}$ hidden layer of the network. First, two average pooling operations in both feature and spatial dimensions is performed to produce the $\mathbf{h}^{(l)} \in \mathbb{R}^{1 \times T \times 1}$, which denotes the average feature value of each skeleton in the sequence. Then, this tensor is introduced to a fully connected layer which transforms features in temporal dimension $T$ to produce the attention tensor $ \mathbf{a}\in \mathbb{R} ^{1 \times T \times 1} = \left \{a_0, a_1, ..., a_T \right \}$ as follows: 
\begin{equation}
\mathbf{a} = Sigmoid\left(\mathbf{h}^{(l)} \mathbf{\Theta}\right),
    \label{eq:tmp-module-fc} 
\end{equation}
where $\mathbf{\Theta} \in \mathbb{R}^{T \times T}$ denotes the learnable transformation matrix, and the resulted attention tensor $\mathbf{a}$ indicates the importance of each skeleton in the sequence. 

To highlight the most informative skeletons, we create the attention tensor $\mathbf{\Lambda} \in \mathbb{R}^{C_{l} \times T \times N}$, a duplicated version of attention map $\mathbf{a}$ with $C_{l} \times N$ copies, which is subsequently dot multiplied to $\mathbf{H}^{(l)}$ as follows: 
\begin{equation}
\mathbf{\hat{H}}^{(l)} = ReLU(\mathbf{H}^{(l)} \otimes \mathbf{\mathbf{\Lambda}}),
    \label{eq:tmp-module-attn} 
\end{equation}
where $\otimes$ denotes element-wise multiplication. 
To select a subset of $T^{\prime}$ skeletons from $\mathbf{\hat{H}^{(l)}}$, the values in the attention map $\mathbf{a}$ are sorted in descending order and the skeletons corresponding to the $T^{\prime}$ highest attention values are selected to be introduced into next layers of the network. 
To improve the computational efficiency of model in both training and inference phases, it would be preferable to use the TAM in the early layers of the network so that the next layers will process less number of skeletons. 

The proposed model is composed of $6$ GCN layers and 1 TAM. The first two GCN layers, map data into $64$ dimensional feature space using spatial convolution followed by batch normalization layer and element-wise ReLU activation function. The mapped data is then introduced into the TAM which selects a subset of skeletons. In order to find the most discriminative skeletons using the TAM, the temporal convolution operation which smoothens the features in temporal domain is not used in the first two GCN layers. 
The $3^{rd}$ and $4^{th}$ layers change data dimensions from $64$ to $128$ and the last two GCN layers increase the number of channels to $256$. In the last $4$ GCN layers, both spatial and temporal convolutions are applied and each operation is followed by batch normalization and element-wise ReLU activation function. All the GCN layers except the first two layers utilize the ResNet \cite{szegedy2017inception} module to benefit from the input skeleton data too. The strides of the temporal convolution layers in $3^{rd}$ and the $5^{th}$ GCN layers are set to $2$ as pooling layer. At the end, the refined spatio-temporal features are introduced into a global average pooling layer which produces an output feature vector of size $256 \times 1$ for each sequence of skeletons and it is introduced into a fully connected layer which is equipped by a SoftMax classifier to classify the action. 
The model is trained with backpropagation in an end-to-end manner to minimize the classification error while it is learning to select the most discriminative skeletons. 

Motivated by the method in \cite{shi2019two}, which utilizes both joints and bones features to enhance the classification performance, we also explore the skeleton bones' length and direction as the second-order information. Each bone is represented as a $3$D vector bounded with two joints. 
The source joint is the one that is closer to the skeletons' center of gravity than the target joint. 
Therefore, each bone pointing from its source joint to the target joint holds both the length and direction information between two joints.

After extracting the bone features from the skeleton data, the two feature tensors are concatenated on their first dimension and the merged joint-bone tensor of size $2C_{in} \times T \times N$ is introduced to the model as input. The overall architecture of the proposed method is shown in Fig. \ref{fig:Model-Diagram}. 

\section{Experiments}
\label{sec:results}
In this section, we describe experiments evaluating the performance of the proposed TA-GCN model in skeleton based human action recognition. We conducted experiments on two widely adopted datasets for evaluating the performance of skeleton-based action recognition methods. These datasets are:

\subsubsection{The \textbf{NTU-RGB+D} \cite{shahroudy2016ntu}} is the largest multi-modality indoor-captured action recognition dataset. The dataset includes RGB videos, infrared videos, $3$D skeletons and depth sequences. The $3$D skeleton data is captured by the Microsoft Kinect-v2 camera and is used in our experiments. This dataset consists of $56,880$ video clips from $60$ different human action classes which are captured from three different views. Each skeleton is represented by $25$ joints which are featured by $3$D coordinate values. In our experiments, we follow exactly the same data splits and benchmark evaluations as in \cite{shahroudy2016ntu}. 
In Cross-View (CV) benchmark, the training set contains $37,920$ samples captured from cameras two and three and the test set contains $18,960$ samples captured from the first camera. 
In Cross-Subject (CS) evaluation, $40,320$ videos which represent $20$ different action classes are used for training and the remaining $16,560$ videos are used for testing. The number of frames for each sample is $300$ and for the samples which have less than $300$ frames, the frame sequence is repeated until it reaches $300$ frames. 
In practice, the input data is a tensor of size $(3 \times 300 \times 25)$.
   
\subsubsection{The \textbf{Kinetics-Skeleton} \cite{kay2017kinetics}} is a very large action recognition dataset that contains $300,000$ video clips of $400$ different human actions collected from YouTube. 
The $2$D joints' coordinates $(x,y)$ on every frame and their confidence score $c$ are estimated using the public OpenPose toolbox \cite{cao2017realtime}. Each skeleton in this dataset contains 18 body joints and each body joint is represented by a 3D vector $(x,y,c)$. 
The number of frames for each sample is fixed to $300$ in a similar way as explained for NTU-RGB+D dataset and the input data would be a tensor of size $(3 \times 300 \times 18)$.
We use the Kinetics skeleton data which is provided by \cite{yan2018spatial}. The training and validation sets contain $240,000$ and $20,000$ skeleton sequences, respectively. 

\subsection{Experimental setting}
All experiments were conducted on PyTorch deep learning framework \cite{paszke2017automatic} with 4 GRX 1080-ti GPUs and batch size of $32$ and $128$ for NTU-RGB+D and Kinetics datasets, respectively. The SGD optimizer is employed to optimize the model with Cross-entropy loss function through back-propagation with weight decay set to $0.0001$. We followed exactly the same setting explained by authors for the baseline method \cite{yan2018spatial} and the state-of-the-art \cite{shi2019two}. In more details, the learning rate is not fixed through all epochs. For NTU-RGB+D dataset, it starts with $0.1$ and it is divided by 10 at epochs $30$, $40$ while the total number of training epochs is fixed to $50$. For Kinetics dataset, it starts with $0.1$ and it is divided by $10$ at epochs $45$, $55$ and the total number of training epochs is fixed to $65$. 
For the Kinetics-Skeleton dataset we don't perform data-augmantation method which is used in \cite{yan2018spatial}. In more details, we utilize all the $300$ skeletons in a sequence.

\subsection{Do we need all skeletons for action recognition?}
\begin{figure}
    \centering
    \includegraphics[width=0.5\textwidth]{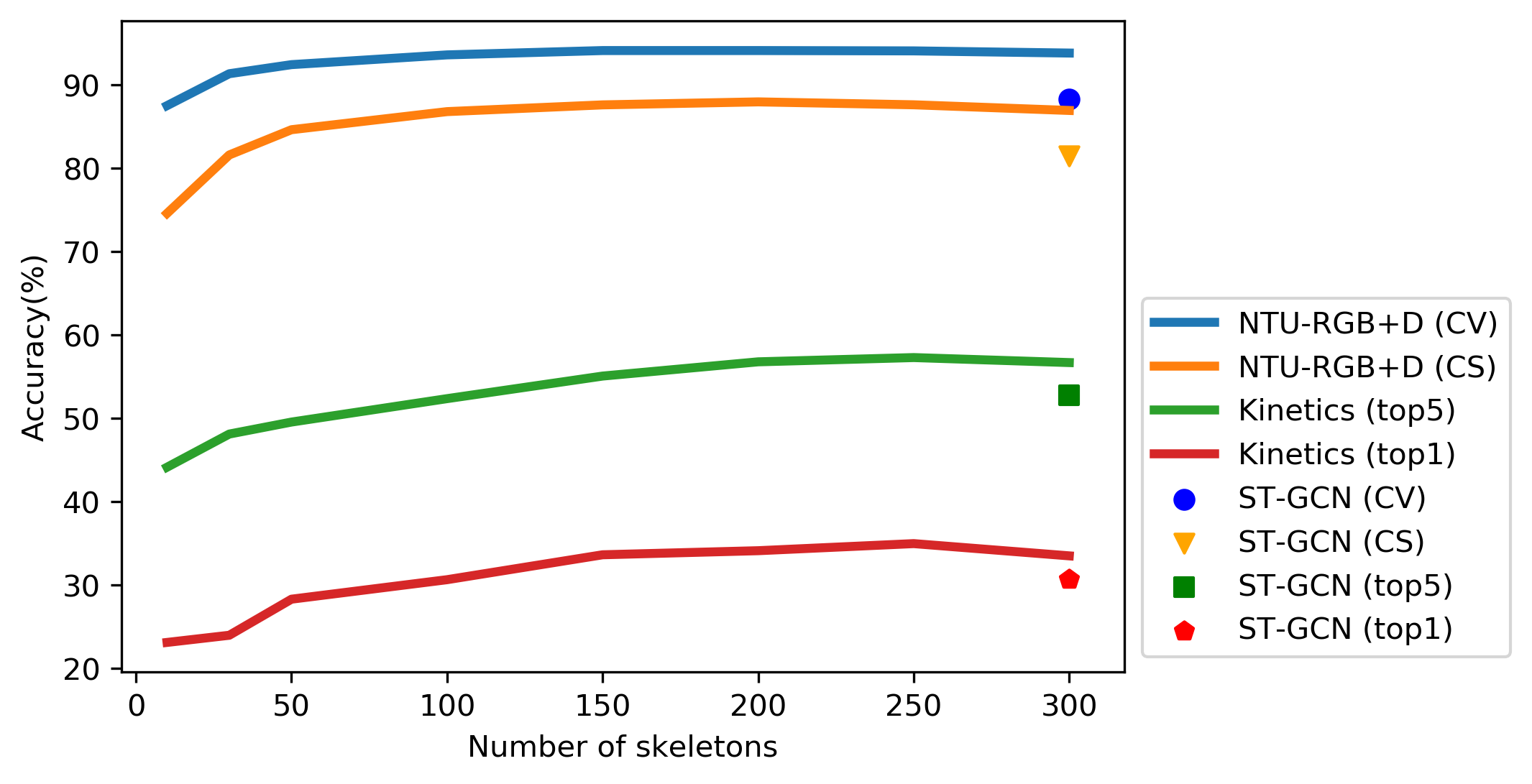}
    \caption{The classification accuracy of the proposed TA-GCN method for a varying number of skeletons ($T^{\prime}$) selected by the proposed TAM. The performance of the baseline ST-GCN is also provided.}
    \label{fig:Skel-ACC}
\end{figure}

The hyperparameter $T^{\prime}$ of the proposed method defines the number of skeletons which are selected by TAM. To evaluate the model's performance with different number of selected skeletons, we applied experiments with varying values of $T^{\prime} = \{ 10,30,50,100,150,200,250,300 \}$. Fig. \ref{fig:Skel-ACC} shows the obtained performance in terms of classification accuracy on Kinetics-Skeletons dataset and NTU-RGB+D dataset with both CV and CS benchmarks. 
As can be seen in Fig. \ref{fig:Skel-ACC}, by increasing the number of skeletons, the models' accuracy is increased gradually. For NTU-RGB+D dataset, the best performance achieved by our method is equal to $94.2\%$ and $87.97\%$ for CV and CS benchmarks, respectively. This performance is achieved by selecting $150$ most informative skeletons and increasing the number of skeletons doesn't improve the performance necessarily. This confirms our hypothesis that the model does not need to process all available skeletons to perform action classification.
Since the Kinetics-Skeleton dataset is more challenging than NTU-RGB+D, both top1 and top5 accuracies are reported. The best top1 and top5 accuracies are $34.95\%$ ad $57.28\%$, respectively, which are achieved by selecting $250$ skeletons.
Overall, we can observe that by applying the proposed TAM for skeleton selection we can achieve competitive performance even for a small number of selected skeletons. For example, the performance in NTU-RGB-D (CV) for $T^{\prime} = 50$ is equal to $92.44\%$, compared to $93.83\%$ corresponding to using all $T=300$ skeletons. This is an advantage for application scenarios with computational restrictions.  

\subsection{Comparison with the state-of-the-art methods}
We compare the performance of the proposed method with the of state-of-the-art methods in Tables. \ref{table:NTU-ACC} and \ref{table:Kinetics-ACC} on the NTU-RGB+D and Kinetics-Skeletons datasets, respectively. 
Table \ref{table:NTU-ACC} is divided in three blocks grouping the methods in three categories, i.e. RNN-based, CNN-based and GCN-based methods, respectively. As can be seen, CNN-based methods perform better than RNN-based methods in general, while the state-of-the-art performance is achieved by GCN-based methods.
\begin{table}[!t]
	\centering
	\caption{Comparisons of the classification accuracy with state-of-the-art methods on the test set of NTU-RGB+D dataset}\footnotesize
	\label{table:NTU-ACC}
	\begin{tabular}{lcccc}
		\hline
		\cline{1-5}
		Method  & CS(\%) & CV(\%) & \#Streams & Skel.sel.\\
		\hline
		HBRNN \cite{du2015hierarchical} & 59.1 & 64.0 & 5 & \xmark \\
		Deep LSTM \cite{shahroudy2016ntu} & 60.7 & 67.3 & 1 & \xmark \\
		ST-LSTM \cite{liu2016spatio} & 69.2 & 77.7 & 1  & \xmark \\
		STA-LSTM \cite{song2017end} & 73.4 & 81.2 & 1  & \cmark \\
		VA-LSTM \cite{zhang2017view} & 79.2 & 87.7 & 1  & \xmark \\
		ARRN-LSTM \cite{li2018skeleton} & 80.7 & 88.8 & 2  & \xmark \\
		\hline
		Two-Stream 3DCNN \cite{liu2017two} & 66.8 & 72.6 & 2  & \xmark \\
		TCN \cite{kim2017interpretable} & 74.3 & 83.1 & 1  & \xmark \\
		Clips+CNN+MTLN \cite{ke2017new} & 79.6 & 84.8 & 1  & \xmark \\
		Synthesized CNN \cite{liu2017enhanced} & 80.0 & 87.2 & 1  & \xmark \\
		3scale ResNet152 \cite{li2017skeleton} & 85.0 & 92.3 & 1  & \xmark \\
		CNN+Motion+Trans \cite{li2017skeletonCNN} & 83.2 & 89.3 & 2  & \xmark \\
		\hline
		ST-GCN \cite{yan2018spatial} & 81.5 & 88.3 & 1  & \xmark \\
		DPRL+GCNN \cite{tang2018deep}  & 83.5 & 89.8 & 1  & \cmark\\ 
		AS-GCN \cite{li2019actional} & 86.8 & 94.2 & 2  & \xmark\\
		2s-AGCN \cite{shi2019two} & 88.5 & 95.1 & 2  & \xmark\\
		GCN-NAS \cite{peng2020learning} & 89.4 & 95.7 & 2  & \xmark\\
		DGNN \cite{shi2019skeleton_directed} & 89.9 & 96.1 & 4  & \xmark\\
		\hline \hline
		\bf{TA-GCN ($T^{\prime} = 150$)} & 87.97 & 94.2 & 1  & \cmark \\ 
		\bf{2s-TA-GCN ($T^{\prime} = 150$)} & 88.5 & 95.1 & 2  & \cmark \\ 
		\bf{4s-TA-GCN ($T^{\prime} = 150$)} & 89.91 & 95.8 & 4  & \cmark \\
		\hline
	\end{tabular}
\end{table}

\begin{table}[!t]
	\centering
	\caption{Comparisons of the classification accuracy with state-of-the-art methods on the test set of Kinetics-Skeleton dataset}\footnotesize
	\label{table:Kinetics-ACC}
	\begin{tabular}{lcccc}
		\hline
		\cline{1-3}
		Method  & Top1(\%) & Top5(\%) & \#Streams & Skel.sel.\\
		\hline
		Deep LSTM \cite{shahroudy2016ntu} & 16.4 & 35.3 & 1 & \xmark\\
		TCN \cite{kim2017interpretable} & 20.3 & 40.0 & 1 & \xmark\\
		ST-GCN \cite{yan2018spatial} & 30.7 & 52.8 & 1 & \xmark\\
		AS-GCN \cite{li2019actional} & 34.8 & 56.5 & 2 & \xmark\\
		2s-AGCN \cite{shi2019two} & 36.1 & 58.7 & 2 & \xmark\\
		DGNN \cite{shi2019skeleton_directed} & 36.9 & 59.6 & 4 & \xmark\\
		GCN-NAS \cite{peng2020learning} & 37.1 & 60.1 & 2 & \xmark\\
		\hline
		1s-TA-GCN ($T^{\prime} = 250$)& 34.95 & 57.28 & 1 & \cmark\\
		2s-TA-GCN ($T^{\prime} = 250$)& 36.1 & 58.72 & 2 & \cmark\\
		4s-TA-GCN ($T^{\prime} = 250$)& 36.9 & 59.77 & 4 & \cmark\\
		\hline
	\end{tabular}
\end{table}

\begin{figure}
    \centering
    \includegraphics[width=0.5\textwidth]{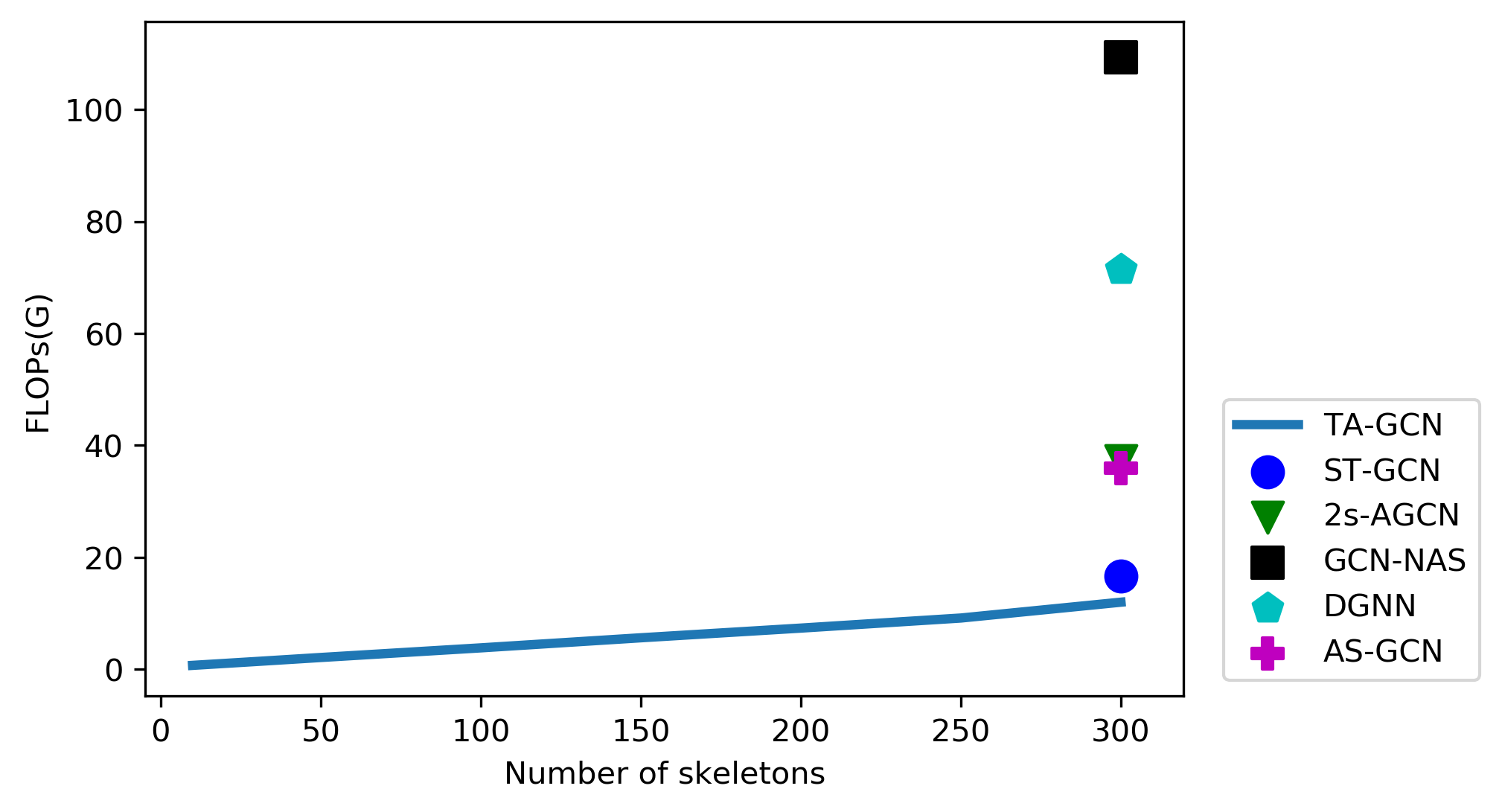}
    \caption{Computational complexity comparison between the proposed method, when it selects different number of skeletons, and all the state-of-the-art methods which utilize all the skeletons of the input sequence.}
    \label{fig:Skel-FLOPs}
\end{figure}

The results for the NTU-RGB+D dataset in Table \ref{table:NTU-ACC} show that the proposed method, TA-GCN, outperforms all the CNN-based and RNN-based methods with a large margin. Besides, TA-GCN outperforms ST-GCN, which is the baseline in GCN-based methods, and DPRL+GCNN by a large margin in both CV and CS benchmarks. Compared to AS-GCN, the proposed method achieves higher accuracy in CS benchmark and a similar performance in CV benchmark. 
Here we should also note that the only competing GCN-based method that performs skeleton selection, i.e. DPRL+GCNN \cite{tang2018deep}, performs poorly compared to all variants of the proposed method.
The TA-GCNs' competitive methods are 2s-AGCN and GCN-NAS and the best performing method is DGNN.
For Kinetics-Skeleton dataset (Table \ref{table:Kinetics-ACC}), the proposed method outperforms the RNN-based methods, ST-GCN and AS-GCN methods and it has competitive performance with 2s-AGCN. The best performing methods on this dataset are GCN-NAS and DGNN.
When the proposed model is trained using two different data streams joints and bones, (2s-TA-GCN), it achieves similar top1 accuracy with 2s-AGCN and its top5 accuracy exceeds 2s-AGCN slightly. The 4s-TA-GCN which is trained using four data streams outperforms DGNN (top5) and achieves competitive performance with GCN-NAS while it has $4.8$ times less computational complexity. 
Table. \ref{table:Params-FLOPs} shows the computational complexity comparison in terms of FLOPs and model parameters (Params) between the GCN-based competing methods on NTU-RGB+D (CV) dataset. Since the source code of the DPRL-GCNN method is not provided by the authors, it is not mentioned in the Table. \ref{table:Params-FLOPs}. 
Fig. \ref{fig:Skel-FLOPs}, illustrates the computational complexity (FLOPs) of our method, when it selects different number of skeletons, and also the state-of-the-art methods which process all the skeletons. It can be seen that TA-GCN has less computational complexity compared to all the competing methods, even when it process all the skeletons in the sequence (i.e. for $T^{\prime}=300$).
\begin{table}[!t]
	\centering
	\caption{Comparisons of the computational complexity with state-of-the-art methods on the on CV benchmark of NTU-RGB+D dataset. The number of FLOPs and Params for 1s-TA-GCN $(T^{\prime} = 150)$ are 5.64G and 2.24M, respectively.}\footnotesize
	\label{table:Params-FLOPs}
	\begin{tabular}{lccc}
		\hline
		\cline{1-3}
		Method  & FLOPs & \# Params \\
		\hline
		1s-TA-GCN $(T^{\prime} = 150)$ & $\times 1$ & $\times 1$ \\
		2s-TA-GCN $(T^{\prime} = 150)$ & $\times 2$ & $\times 2$ \\
		4s-TA-GCN $(T^{\prime} = 150)$ & $\times 4$ & $\times 4$ \\
		ST-GCN & $\times 2.9$ & $\times 1.3$ \\
		AS-GCN & $\times 6.3$ & $\times 3.2$ \\
		2s-AGCN & $\times 6.6$ & $\times 3$ \\
		DGNN & $\times 12.6$ & $\times 3.6$ \\
		GCN-NAS & $\times 19.3$ & $\times 8.9$ \\
		\hline
	\end{tabular}
\end{table}

Columns $3$ and $4$ in Tables \ref{table:NTU-ACC} and \ref{table:Kinetics-ACC} indicate the number of network streams used by each GCN-based method and whether each method performs skeleton selection (or otherwise employs all $T=300$ skeletons). As can be seen, methods employing more than one streams generally outperform those with one stream. 
Although these methods achieve higher classification accuracy compared to single-stream methods, they have high computational complexity. 
Similar to our proposed method, the ST-GCN and DPRL-GCNN methods train only one network stream. They utilize only the joint data to train the model and we fuse both joint and bone information in the first layer (input) which prevents increasing number of model parameters and FLOPs. 
The results show that not only does our method outperform the baseline ST-GCN in terms of classification accuracy, but also it has 2.9 times less FLOPs and 1.3 times less number of parameters. Therefore, TA-GCN can be a strong and efficient baseline for GCN-based human action recognition which achieves better performance than ST-GCN. 
The 2s-AGCN, GCN-NAS methods are built on top of ST-GCN method and train two networks using joint and bone data separately which doubles the number of model parameters and FLOPs. 
DGNN which has the best classification accuracy on both datasets trains 4 different networks with joint, bone, joint-motion and bone-motion data.
DGNN and GCN-NAS have 12.6 and 19.3 times more FLOPs than the proposed method, respectively, and from 3.6 to 8.9 times more parameters.

To conduct a fair comparison, we also trained TA-GCN method with 4 different data streams separately. In Table. \ref{table:NTU-ACC}, \ref{table:Kinetics-ACC}, 2s-TA-GCN shows the ensembled softmax scores of the trained models with joint and bone data and 4s-TA-GCN shows the classification accuracy of the model when the softmax scores of all the 4 streams, joint, bone, joint-motion and bone-motion, are ensembled. The results indicate that training multiple models and ensembling the softmax scores can increase the accuracy by at least $1\%$. 2s-TA-GCN has a similar accuracy with 2s-AGCN while it has 3.3 times less FLOPs and 1.5 times less parameters. 4s-TA-GCN outperforms the competing methods 2s-AGCN and GCN-NAS with 1.65, 4.82 times less FLOPs, respectively. 
As can be seen in Table. \ref{table:Params-FLOPs}, GCN-NAS has the maximum number of FLOPs and it has competitive performance with 2s-TA-GCN which has with 9.6 times less computational complexity.  
In comparison with state-of-the-art, our proposed TA-GCN achieves good performance with much less computational complexity which makes TA-GCN applicable for many practical tasks with limited computational capacity. 

\section{Conclusion}
\label{sec:conclusion}
In this paper, we proposed a temporal attention-augmented GCN to improve the computational efficiency in skeleton-based action recognition. Our method trains the attention mechanism to select the most informative skeletons for each action in an end-to-end manner. On two widely used benchmark datasets, the proposed method has competitive performance with the state-of-the-art, while being up to 10 times less computationally complex, and it outperforms the baseline in terms of both classification accuracy and computational complexity with a large margin. Therefore, it can be an efficient and strong baseline for skeleton-based action recognition.

\section*{Acknowledgment}
This work received funding from the European Union’s Horizon 2020 research and innovation programme under grant agreement No 871449 (OpenDR). This publication reflects the authors’ views only. The European Commission is not responsible for any use that may be made of the information it contains.

\bibliographystyle{IEEEtran}
\bibliography{root}

\end{document}